\documentclass[10pt,twocolumn,letterpaper]{article}

\makeatletter
\@namedef{ver@everyshi.sty}{}
\makeatother

\newif\ifarxiv
\arxivtrue 

\ifarxiv
  \usepackage[pagenumbers]{cvpr}
  \newcommand{\ARXIVversion}[2]{#1}
\else
  \usepackage{cvpr}
  \errmessage{Update ARXIV paper number in the bib!}
  \newcommand{\ARXIVversion}[2]{#2}
\fi

\usepackage{graphicx}
\usepackage{amsmath}
\usepackage{amssymb}
\usepackage{booktabs}

\usepackage[pagebackref,breaklinks,colorlinks]{hyperref}

\usepackage[capitalize]{cleveref}
\crefname{section}{Sec.}{Secs.}
\Crefname{section}{Section}{Sections}
\Crefname{table}{Table}{Tables}
\crefname{table}{Tab.}{Tabs.}

\usepackage{todonotes}
\usepackage{amsmath}
\usepackage{amssymb}
\usepackage{stmaryrd}
\usepackage{graphicx}

\usepackage{multirow}
\usepackage{multicol}
\usepackage{pifont}
\usepackage{makecell}
\usepackage{algorithm}
\usepackage{algpseudocode}
\usepackage{caption}
\usepackage{flushend}

\DeclareMathOperator*{\argmax}{arg\,max}

\newcommand{\paragraphcustom}[1]{\vspace{3pt}\noindent\textbf{#1}}
\newcommand{\paragraphcustomWOvspace}[1]{\noindent\textbf{#1}}

\usepackage[normalem]{ulem}

\newcommand{\cmark}{\ding{51}}%
\newcommand{\xmark}{\ding{55}}%

\makeatletter
\newcommand\footnoteref[1]{\protected@xdef\@thefnmark{\ref{#1}}\@footnotemark}
\makeatother

\def\cvprPaperID{1784} 
\def\confName{CVPR}
\def\confYear{2023}

\begin{document}

\title{Multi-Task Learning of Object State Changes from Uncurated Videos}

\author{
	Tom\'{a}\v{s} Sou\v{c}ek\textsuperscript{1}
	\quad\quad\quad
	Jean-Baptiste Alayrac\textsuperscript{2}
	\quad\quad\quad
	Antoine Miech\textsuperscript{2}
	\\
	Ivan Laptev\textsuperscript{3}
	\quad\quad\quad
	Josef Sivic\textsuperscript{1}
	\\
	\small{$^1$CIIRC CTU \quad $^2$DeepMind \quad $^3$ENS/Inria}
	\\
	\small{\texttt{tomas.soucek@cvut.cz}}
	\\
	\small{\url{https://soczech.github.io/multi-task-object-states/}}
}

\twocolumn[{
\maketitle

\vspace{-1.2em}
\vspace{-0.5em}
\centering
\includegraphics[width=1.0\linewidth]{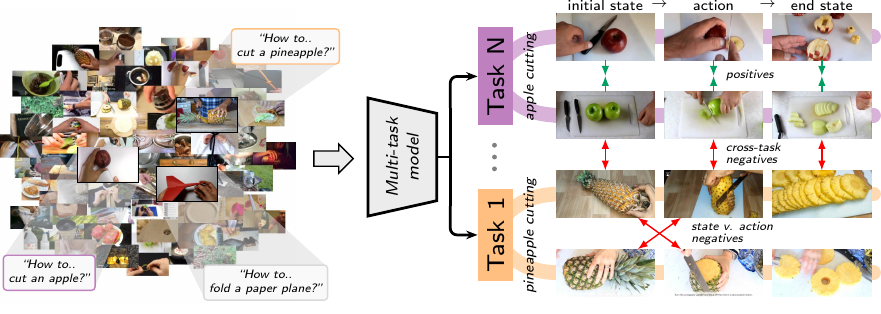}\vspace{-0.5em}
\captionof{figure}{{\bf Multi-task self-supervised temporal localization of object states and actions in videos.} 
Our approach takes as input a collection of uncurated web videos obtained by searching the Internet for queries such as \textit{``How to cut an apple?"} or \textit{``How to cut a pineapple?"} (left) and imposes constraints between the different recognition tasks (here  \textit{apple cutting} and \textit{pineapple cutting}) in order to temporally localize the initial object state (\eg \textit{whole apple}), the state-modifying action (\eg \textit{cutting apple}), and the end state (\eg \textit{sliced apple}) in a video.
}
\label{fig:teaser}
\vspace{1.4em}
}]

\begin{abstract}
We aim to learn to temporally localize object state changes and the corresponding state-modifying actions by observing people interacting with objects in long uncurated web videos.
We introduce three principal contributions. First, we explore alternative multi-task network architectures and identify a model that enables efficient joint learning of multiple object states and actions such as pouring water and pouring coffee. Second, we design a multi-task self-supervised learning procedure that exploits different types of constraints between objects and state-modifying actions enabling end-to-end training of a model for temporal localization of object states and actions in videos from only noisy video-level supervision.
Third, we report results on the large-scale ChangeIt and COIN datasets containing tens of thousands of long (un)curated web videos depicting various interactions such as hole drilling, cream whisking, or paper plane folding. We show that our multi-task model achieves a relative improvement of 40\% over the prior single-task methods and significantly outperforms both image-based and video-based zero-shot models for this problem. We also test our method on long egocentric videos of the EPIC-KITCHENS and the Ego4D datasets in a zero-shot setup demonstrating the robustness of our learned model.
\end{abstract}


\footnotetext[1]{Czech Institute of Informatics, Robotics and Cybernetics at the Czech Technical University in Prague.}
\footnotetext[3]{D\'{e}partement d'informatique de l'ENS, \'{E}cole normale sup\'{e}rieure, CNRS, PSL Research University, 75005 Paris, France.}

\section{Introduction}
Learning changes of object states~\cite{Brady06} together with the corresponding state-modifying actions is an important problem in embodied video understanding. 
While a supervised approach to this problem could be an option, the huge variety of existing object states prohibits the use of manual image and video annotation. 
To address this issue, previous works demonstrate successful learning of object states from (un)curated web videos by observing person-object interactions~\cite{Alayrac16ObjectStates,soucek2022lookforthechange}.
These methods, however, learn independent models for each type of interaction, hence, scaling them to many states and actions becomes impractical.
Moreover, \cite{Alayrac16ObjectStates,soucek2022lookforthechange} ignore relations among different interactions as, for example, {\em pouring coffee} and {\em pouring water} into a cup both change the state of a cup from {\em empty} to {\em full}. 

In this work, we address the above limitations and introduce a multi-task learning approach that {\em jointly} learns state changes and state-modifying actions from {\em multiple} types of human-object interactions.
Toward this goal, we aim to answer the following questions:~(i) what is the appropriate multi-task model for effective sharing of information across related tasks  such as \textit{cutting apple} and \textit{cutting pineapple}?~And (ii) how to learn such a model to enable temporal localization of object states and state-modifying actions from long uncurated videos automatically obtained by querying a search engine for queries such as \textit{``How to cut a pineapple"}?
To answer these questions, we describe the following contributions.
First, we explore alternative multi-task network architectures and identify a model that enables efficient {\em joint} learning of multiple object states and actions.
Second, we design a multi-task self-supervised learning procedure that exploits different types of constraints between objects and state-modifying actions. 
For example, these constraints enforce that no action is predicted when an object (such as \textit{an apple}) is predicted to be in one of its states (\ie{} \textit{whole} or \textit{sliced}), or that the video depicts only one type of state-modifying action. Using these constraints, we learn a model for temporal localization of object states and actions in videos from only noisy video-level supervision provided by querying an Internet search engine.
Third, we perform experiments on four different datasets and show that our multi-task model outperforms prior work as well as both image-based and video-based zero-shot methods.
In particular, our approach outperforms prior methods by the relative improvement of 40\% on the challenging ChangeIt dataset~\cite{soucek2022lookforthechange}
and achieves state-of-the-art results for unsupervised action discovery on the COIN dataset~\cite{tang2019coin}.
Finally, we investigate the performance of our model in zero-shot setup on egocentric videos and demonstrate that our model trained on noisy web videos can be successfully applied to the EPIC-KITCHENS~\cite{damen2018scaling} and Ego4D~\cite{grauman2021ego4d} datasets.

\section{Related work}

\paragraphcustomWOvspace{Learning object states from videos.} 
Detecting object states in images~\cite{misra2017red,nagarajan2018attributes,purushwalkam2019task,yang2020learning,saini2022disentangling} or learning actions~\cite{saini2022recognizing} and their modifiers~\cite{doughty2020action,doughty2022you} in videos has been addressed in supervised settings.
Others have shown that the causality of actions can be leveraged as an additional supervisory signal~\cite{fernando2015modeling,wei2018learning}.
Interestingly, object states in videos can be implicitly discovered by predicting what happens in the future in the video~\cite{epstein2021learning,jayaraman2018time}. In egocentric videos, detected object bounding boxes can be assigned with object states~\cite{liu2017jointly}. More recently, some of the goals of the Ego4D benchmark~\cite{grauman2021ego4d} are to regress a bounding box of a manipulated object, detect a point of no return during a person-object interaction (e.g. when the object is lifted from the table), or classify whether a state change happened in a short few-second-long clip. In contrast, our work addresses a very different problem of discovering the temporal locations of object states and the corresponding state-modifying actions in long uncurated web videos in a self-supervised manner without temporal annotations of object states or actions. 
We review related methods for this problem next.

\paragraphcustom{Self-supervised object state and action learning.}
The problem of self-supervised object state and action learning has been studied previously but in a small-scale curated setup~\cite{Alayrac16ObjectStates} or with multiple views for a single scene~\cite{damen2014you}. 
Sou\v{c}ek \etal{}~\cite{soucek2022lookforthechange} recently explored this problem at a larger scale by learning from uncurated Internet videos.
The main idea is to use an EM-style approach that alternates between \textbf{(i)}~localizing the actions and object states given the current model predictions and constraints derived from assumptions about the problem (\eg~the action should be in between the two states) and \textbf{(ii)}~using these locations to compute temporal pseudo labels which are then used to update the weights of the object state and action classifiers using a standard cross-entropy loss. 
We build on this approach by using a similar EM-style learning method while addressing a fundamental limitation of this approach.
Indeed, while \cite{soucek2022lookforthechange} requires training a separate model for each category, we instead introduce a multi-task approach that demonstrates significant quantitative and qualitative improvements as well as an improved ability to scale to many videos depicting a larger variety of tasks.

\paragraphcustom{Unsupervised temporal action segmentation.} Another line of work addressed the problem of unsupervised or self-supervised temporal segmentation of long tasks composed of multiple sub-actions, such as \textit{making a salad}. Some of these methods utilize their model's predictions in a Viterbi algorithm to compute pseudo labels for training the model~\cite{li2019weakly,richard2018neuralnetwork,soucek2022lookforthechange} with the possibility to substitute Viterbi by its differentiable variant~\cite{chang2019d3tw}. Others pre-train a feature extractor in a self-supervised manner and then use clustering and additional post-processing to discover the sub-actions~\cite{kukleva2019unsupervised, wang2022sscap}.
Discovered actions from the clustering can be used as pseudo labels to train better action embeddings \cite{li2021action}. Pseudo labels are also used to contrast action instance features with background features to produce action scores for each video segment \cite{lee2021learning}.
Others temporally segment actions in videos using weak alignment between text and video \cite{shen2021learning}, using task hierarchies \cite{ghoddoosian2022hierarchical}, or by parsing videos into so-called activity threads~\cite{price2022unweavenet}. Finally, fully-supervised learning can be used to train a model predicting seen objects/actions but in unseen combinations~\cite{materzynska2020something}.
In contrast, we work with significantly larger, possibly uncurated datasets only with noisy video-level labels and focus on the relationship between actions and objects rather than finding sub-actions of a specific task.

\paragraphcustom{Multi-task learning.}
Previous work on state and action discovery trains a separate model for each task~\cite{soucek2022lookforthechange}. 
Yet there have been many successful applications of multi-task learning where a single model trained on many tasks jointly can benefit from the additional related training data for the other tasks, as detailed next.
Cross-task data leads to more general representations in natural language processing \cite{liu2019multi,subramanian2018learning}, computer vision \cite{liu2019end,misra2016cross,simonyan2014two,taskonomy2018,zhukov2019cross} as well as in combined vision-language tasks \cite{hu2021unit}. 
A similar situation is in self-supervised training where it has been shown that combining multiple tasks also improves learned features compared to single-task baselines~\cite{doersch2017multi,ren2018cross}.
Further, vision-language models using contrastive training~\cite{align,miech2020end,clip,zellers2022merlot,xu2021videoclip} on large-scale datasets exhibit zero-shot multi-task generalization with some models also directly focused on object state changes and egocentric videos~\cite{lin2022egocentric,mittal2022learning}.
It has also been shown that multi-task learning improves performance for the closely related problem of localizing key steps in instructional videos, even in self- or weakly-supervised setups \cite{elhamifar2020self,zhukov2019cross}. 
To the best of our knowledge, we are the first to successfully apply parameter sharing \cite{caruana1997multitask} to the problem of self-supervised object state and action discovery. This enables information sharing between similar interaction categories such as \textit{pouring water} and \textit{pouring coffee} and generates more useful feature representations for each task resulting in significant gains in prediction accuracy.

\section{Method}
We are given a set of uncurated videos likely to depict an object state change associated with a given interaction category.
For example, an interaction category could be \textit{box wrapping} which changes the state of a \textit{box} from \textit{unwrapped} to \textit{wrapped}.
These videos are uncurated as they are obtained from searching the web~\cite{soucek2022lookforthechange} with a given query, \eg{} ``\textit{how to wrap a box}''.
As a result, there are no guarantees that the action will happen in all videos.
This allows to properly test algorithms in realistic scenarios with the prospect of learning about human actions at scale.
Given this noisy video collection, our goal is to learn a model able to temporally localize the actions as well as to identify frames where the object appears in either one of the states (pre- and post-action).
One important challenge is that we are not given any temporal annotations of actions nor states.
Instead, we seek to infer this information based solely on the noisy video-level category labels provided by the Internet search results.

To address this challenge, prior works~\cite{Alayrac16ObjectStates,soucek2022lookforthechange} leverage the assumption that the state-modifying action must temporally succeed the object in its initial state and precede the object in its end state.
However, the learning process is performed independently for all interaction categories, hence not leveraging the potential of shared information across interaction categories, \eg~between \textit{apple cutting} and \textit{pineapple cutting}.
We build on this prior work but instead model all interaction categories jointly
and demonstrate that this greatly improves performance.
This multi-task setup relies on careful model design choices that allow sharing parameters across interaction categories (Section~\ref{sec:model_architecture}) as well as novel label constraints (Sec.~\ref{ref:model_constraints}), which help obtain better self-supervision for our problem and also help train our vision backbone end-to-end for improved performance.

\subsection{Detecting object states and state-modifying actions in a multi-task setup}
\label{sec:model_architecture}
In this section, we investigate several multi-task architectures for detecting object states and state-modifying actions in videos. The key open question is how to best leverage the fact that the different tasks are related. For example, \textit{egg peeling} and \textit{egg whisking} both consider the same object. In addition, when the model is confident that a particular video segment contains the action \textit{whisking}, this segment is unlikely also to contain the action \textit{peeling} and hence can serve as a negative example for learning the \textit{peeling} action detector. Therefore, the classifiers for the different object states and actions could interact both \textbf{(i)} by sharing parameters to exploit the fact that objects and actions in the world are related, and \textbf{(ii)} at the level of the outputs of the model to leverage the fact that a particular video segment can only contain a single action or object state change. We formalize these notions next.

\begin{figure}[t]
    \centering
    \includegraphics[width=0.99\linewidth]{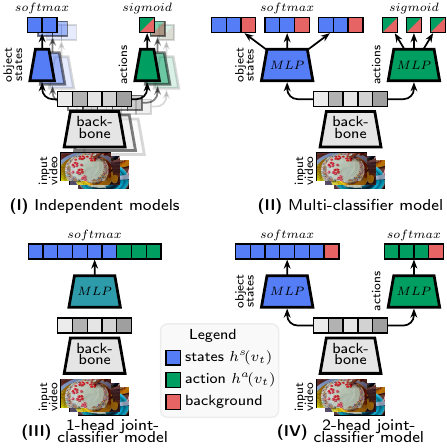}
    \vspace{-0.5em}
    \vspace{-0.2em}
    \caption{
    \textbf{Multi-task architectures for learning object states and actions from videos.} See Section~\ref{sec:model_architecture} for details.
    }
    \label{fig:architectures}
    \vspace{-0.5em}
    \vspace{-0.5em}
    \vspace{-0.5em}
\end{figure}

\paragraphcustom{Multi-task model architectures.} 
More formally, we introduce a {\em single model} $h$ that given a frame of a video $v_t$ provides a confidence score for the initial state $h_c^{s_1}(v_t)$, the end state $h_c^{s_2}(v_t)$, and the action $h_c^{a}(v_t)$ corresponding to an interaction category $c$. For example, for the category \textit{box wrapping}, the initial state $s_1$ is \textit{unwrapped box}, the end state  $s_2$ is \textit{wrapped box}, and the action $a$ is \textit{wrapping}.
We investigate two main architecture design choices. First, we investigate whether to have a single head that predicts both actions and object states together, or whether to predict actions and object states separately. Second, we investigate whether to predict independent scores using multiple independent classifiers, one for each of the tasks (\eg{} \textit{apple cutting} and \textit{pineapple cutting}), or to predict all tasks jointly using a single joint classifier.
In particular, we investigate the following four architectures, shown in Figure \ref{fig:architectures}, and discuss their benefits and drawbacks next.

\textbf{I.} \textit{Independent models.} The first option, employed by the prior work \cite{soucek2022lookforthechange} and included for reference, is a set of independent single-task classifiers. It requires a separate model to be trained for each task and does not allow sharing of information between the individual interaction categories. 
Each single-task classifier outputs for each video frame $v_t$ two logits normalized by a softmax function to produce the likelihood of the initial and the end state. Also, each classifier outputs the likelihood of action, computed as a sigmoid of the single output logit.

\textbf{II.} \textit{Multi-classifier model.} A common approach to multi-task learning is to have a separate classification head for each task.
Unlike option \textbf{I.}, this architecture allows for information sharing between different interaction categories but the state and action classification of video frames is done independently for each category. 

\textbf{III.} \textit{1-head joint-classifier model.} A simple option for a multi-task classifier is to predict a distribution over all classes using a softmax function.
In our case, given $N$ interaction categories such as \textit{box wrapping}, we predict a distribution over $3N$ possible outputs -- for each category, it is the likelihood of the action, the initial and the end state. 

\textbf{IV.} \textit{2-head joint-classifier model.}
Finally, we can 
disentangle the state and action predictions into two classifiers, as shown in Figure \ref{fig:architectures} (IV). The action head predicts distribution over actions from all $N$ categories and one background (\textit{no-action}) class. Similarly, the state head predicts distribution over the initial and the end states for all categories, together $2N$ outputs or $2N+1$ with the optional background (\textit{no-state}) class.
The separate state and action classifiers allow a video frame to be classified as both the action and some of the two states, if needed, for example, in case of long-duration actions such as \textit{bacon frying} (the bacon stays some time in the initial state, \ie{} \textit{raw}, at the beginning of \textit{frying}).
Importantly, during training, it also allows the use of two pseudo labels for a single video frame, one action label and one state label, as described in Section \ref{ref:model_constraints}.

\paragraphcustom{Model inference in the multi-task setup.}
Given a video $v$ and its category $c$ such as \textit{box wrapping}, the previous work \cite{soucek2022lookforthechange} temporally localizes the corresponding action and the object in the initial and in the end state in the video (\eg{} \textit{wrapping}, \textit{unwrapped box}, and \textit{wrapped box}). However,~\cite{soucek2022lookforthechange} requires the interaction category $c$ of the video to be known, which limits the applicability of the method. Our multi-task approach addresses this limitation and can be applied to new videos where the interaction category is unknown. 
In detail, our model $h$ produces the score of the action $h_c^{a}(v_t)$, the initial state $h_c^{s_1}(v_t)$, and the end state $h_c^{s_2}(v_t)$ corresponding to the category $c$ for every video frame $v_t$ and every category $c$. We use our model predictions with the causal ordering constraint~\cite{Alayrac16ObjectStates, soucek2022lookforthechange} to compute the score $p(c|v)$ that a given video $v$ contains an object change of category $c$ as follows:
\begin{align}
	p(c|v) \stackrel{\text{def}}{=} \max_{i,j,k \in \mathcal C} \ & h_c^{s_1}(v_i)\cdot h_c^{a}(v_j) \cdot h_c^{s_2}(v_k),
	\label{eq:max}
\end{align}
where $\mathcal C$ implements the causal ordering constraints, \ie{} that the location of the action has to be after the initial state but before the end state.
For the purpose of classification, we use $\hat{c}=\argmax_c p(c|v)$ as the most probable category $\hat{c}$ for video $v$. Then, we can use the same constraint to compute the temporal locations of the initial state $d_{\hat{c}}^{s_1}(v)$, the end state $d_{\hat{c}}^{s_2}(v)$ and the action $d_{\hat{c}}^{a}(v)$ for the computed category $\hat{c}$ as follows:
\begin{align}
        d_{\hat{c}}^{s_1},\ d_{\hat{c}}^{a},\ d_{\hat{c}}^{s_2} = \argmax_{i,j,k \in \mathcal C} \ & h_{\hat{c}}^{s_1}(v_i)\cdot h_{\hat{c}}^{a}(v_j) \cdot h_{\hat{c}}^{s_2}(v_k).
	\label{eq:argmax}
\end{align}
A single video can contain multiple categories of object changes; thus, the scores $p(c|v)$ for all categories $c$ need not to sum to one. For the same reason, we can use Equation~\eqref{eq:argmax} for categories with the score $p(c|v)$ higher than some threshold to detect multiple categories in a single video.

\subsection{Multi-task self-supervised training with pseudo labels}
\label{ref:model_constraints}

Our objective is to train the multi-task model $h$ for object state and action localization introduced in Section \ref{sec:model_architecture}. 
We seek to train the model with a set of uncurated videos with only noisy {\em video-level} category labels obtained by searching for the interaction category (\eg{} \textit{box wrapping}) using a video search engine~\cite{soucek2022lookforthechange}. 
In particular, these videos do not have any temporal annotations of action or object state locations, which are difficult and expensive to obtain at scale.
The main challenge in the multi-task setup is how to best leverage the complementary nature of object states and actions together with the noisy video-level labels in order to improve upon the single-task model~\cite{soucek2022lookforthechange}.

\paragraphcustom{Multi-task self-supervised learning algorithm.} 
We address this challenge by a multi-task self-supervised algorithm, where the object state and action classifiers learn from each other's localization outputs in an analogy to multi-modal self-supervised learning~\cite{Alayrac2020MMVN} where the classifier for one modality (\eg{} audio) learns from the outputs of classifiers for other modalities (\eg{} video or language). In our case, we use such {\em localization self-supervision} together with noisy video-level labels to learn a model capable of localizing object states and actions in videos. Our approach works in a multi-task setup where classifiers for several actions and object states are learned jointly thus providing an additional (self-)supervisory signal to each other. Our approach operates in three steps. First, given the current model $h$ we predict the most probable locations of actions and object states for each training video. Second, we construct from these predictions a set of temporal {\em pseudo labels}~$\mathcal T$ capturing various types of constraints between the action and object state localization tasks. Finally, we update the parameters of the model $h$ using the constructed set of pseudo labels $\mathcal T$ using a standard back-propagation step. This three-step algorithm is summarized in Algorithm~\ref{alg} and detailed below.

\begin{algorithm}[t]
\caption{Multi-task learning}\label{alg}
\begin{algorithmic}
\Repeat
    \State sample batch $\mathcal{B}$ of videos $v$ and their categories $l$
    \State $\mathcal{T} \gets \emptyset$
    \For{$v,l\in\mathcal{B}$}
\State \(\triangleright\) \textit{\textbf{Step 1:} Detect states and actions,  Eq. \eqref{eq:argmax}}
        \State $d_{l}^{s_1}$, $d_{l}^{s_2}$, $d_{l}^{a} \gets$ \Call{Detect}{$v$,$l$}
\State \(\triangleright\) \textit{\textbf{Step 2:} Compute labels w/ constraints, Sec. \ref{ref:model_constraints}}
        \State $\mathcal{T} \gets \mathcal{T} \cup$ \Call{PseudoLabels}{$v$, $d_{l}^{s_1}$, $d_{l}^{s_2}$, $d_{l}^{a}$} 
    \EndFor
    \State update model $h$ by $\nabla_h \mathcal{L}(\mathcal{T})$ 
    \Comment{\textit{\textbf{Step 3:} Train, Eq. \eqref{eq:loss}}}
\Until{convergence}
\end{algorithmic}
\end{algorithm}

Formally, we are given a set of videos $v$ with their noisy video-level interaction category labels~$l$. The goal is to learn the model $h$ that predicts the likelihood of the initial state $h_c^{s_1}(v_t)$, the end state $h_c^{s_2}(v_t)$, and the action $h_c^{a}(v_t)$ for every video frame $v_t$ and every category $c$.
In detail, the three steps of the algorithm outlined above are as follows.
{\bf Step~1:} Sample a batch of videos $v$ with their labels~$l$ and for each video detect the locations of the initial state $d_l^{s_1}(v)$, the end state $d_l^{s_2}(v)$, and the action $d_l^{a}(v)$ using Eq.~\eqref{eq:argmax}, Section~\ref{sec:model_architecture}. {\bf Step~2:} Use these inferred locations to compute a set of temporal pseudo labels $(v,t,c,q)\in\mathcal{T}$ which define the desired classifier output $h_c^q(v_t)$ (\eg{} $q=s_2=$ \textit{wrapped box} for category $c=$ \textit{box wrapping}) for the video segment~$v_t$. {\bf Step~3:} The model weights are updated using the gradients of the cross entropy function $\mathcal{L}(\mathcal{T})$ (outlined below in Eq.~\eqref{eq:loss}). 
Next, we provide details of the loss function $\mathcal{L}(\mathcal{T})$ and the set of pseudo-labels $\mathcal T$.

\paragraphcustom{Multi-task loss function.} In our algorithm, we take a gradient step according to the following loss
\begin{align}
    \mathcal{L}(\mathcal{T})=-\sum_{v,t,c,q\in\mathcal{T}}w(v,q)\log h_c^q(v_t),\label{eq:loss}
\end{align}
where $w(v,q)$ is a scalar weight determining the per-video contribution to the loss. 
In more detail, we use the video-level noise adaptive weight introduced in \cite{soucek2022lookforthechange}, which gives a confidence score about the relevance of the video $v$.
The term $\log h_c^q(v_t)$ is the cross-entropy between the output of the classifier $h$ for frame $v_t$ and the pseudo-label $q$.
For each category $c$, the classifier predictions include the initial state ($q=s_1$), the end state ($q=s_2$), the action ($q=a$), and the background classes $q=\emptyset_s$ and $q=\emptyset_a$ for the states and action, respectively.

\paragraphcustom{Constructing the set of pseudo labels.} Correctly designing the set of pseudo labels $\mathcal{T}$ is crucial for training a well-performing model as the pseudo labels are the only supervision for the model.
We base the construction of this set $\mathcal{T}$ on multiple assumptions detailed next. 

\begin{figure}[t]
    \centering
    \includegraphics[width=\linewidth]{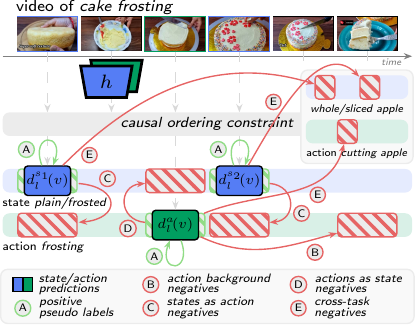}
    \vspace{-0.5em}
    \vspace{-0.5em}
    \vspace{-0.5em}
    \caption{
    {\bf Constructing pseudo labels $\mathcal T$} for training our multi-task object state and action recognition model $h$.
    The pseudo labels (dashed boxes) are inferred from the predicted state and action locations $d_l(v)$ using a set of constraints (indicated by the arrows). While the pseudo labels (A) and (B) have been used in prior single-task set-up~\cite{soucek2022lookforthechange}, we introduce new forms of supervision based on the following premises: no action should be predicted at the locations of states (C), no state should be predicted at the locations of actions (D), and states (or actions) of different categories should serve each other as negatives (E). This additional supervision significantly improves the final classifier's performance. 
    }
    \label{fig:targets}
    \vspace{-0.5em}
    \vspace{-0.5em}
\end{figure}

\textbf{(A)} \textit{Positive action and state pseudo labels} \cite{soucek2022lookforthechange}. Given a video $v$, its category label $l$ and associated locations of the initial state $d_l^{s_1}(v)$, end state $d_l^{s_2}(v)$ and action $d_l^{a}(v)$, we follow the prior work \cite{soucek2022lookforthechange} and first construct positive pseudo labels reinforcing the network $h$ in predicting the respective states and action at their inferred locations. 
We illustrate these positive pseudo-labels in Figure~\ref{fig:targets} as A. 
Formally, for the initial state, we add the quadruple $(v,t,l,s_1)$ to the set $\mathcal{T}$ for every frame $t$ close enough to $d_l^{s_1}(v)$, \ie{} $|t-d_l^{s_1}(v)|\leq \delta$, where $\delta$ is a parameter. 
We proceed similarly with the end state and the action. 

\textbf{(B)} \textit{Action background negatives} \cite{soucek2022lookforthechange}.
Non-action (action background) pseudo labels are generated at a fixed distance $\delta'$ from the predicted action location $d_l^{a}(v)$. Formally, the quadruple $(v,t,l,\emptyset_a)$ is added for every frame $t$ at a certain distance range given by the parameter $\delta$, \ie{} $\delta' \leq |t-d_l^{a}(v)|\leq \delta' + \delta$   (illustrated as B in Figure~\ref{fig:targets}). 

\textbf{(C)} \textit{States as action negatives.} 
We notice that the action classifier sometimes struggles to differentiate the states from the action. 
For instance, a frame showing an unlit candle is selected as the action \textit{candle lighting} instead of the frame where the person lights the candle. 
To address this, we enforce the action classifier to predict the action background label $\emptyset_a$ at the predicted initial and end state locations $d_l^{s_1}(v)$, $d_l^{s_2}(v)$ (illustrated in Figure \ref{fig:targets} as C). 
Formally, the quadruple $(v,t,l,\emptyset_a)$ is added to the set $\mathcal{T}$ for every frame $t$ satisfying $|t-d_l^{s_1}(v)|\leq \delta$ or $|t-d_l^{s_2}(v)|\leq \delta$.

\textbf{(D)} \textit{Actions as state negatives.} For some objects, such as cream in a bowl, it may be easier to identify them by an accompanying tool such as a whisk. However, in our scenario, this behavior is undesired as the presence of a tool is more likely to indicate an action. Therefore, we add a background class for the state classifier and enforce the state classifier to predict the background state class $\emptyset_s$ at the predicted action location $d_l^a(v)$. 
Formally, we add the quadruple $(v,t,l,\emptyset_s)$ for every frame $t$ satisfying $|t-d_l^{a}(v)|\leq \delta$.
This is illustrated in Figure~\ref{fig:targets} as D.

\textbf{(E)} \textit{Cross-task negatives.} Until now, we considered only in-video pseudo labels, \ie{} constructed for video $v$ using classifier's prediction for the same video $v$. If a model is shown a video without the object it has been trained to localize, its predictions could be random. This is also one of the main weaknesses of the prior work \cite{Alayrac16ObjectStates, soucek2022lookforthechange}. In our multi-task training, however, we can utilize the other video categories in a batch to mine negatives for the trained classifiers. For our architectures with softmax over all categories (Section \ref{sec:model_architecture}, option {III.} and {IV.}), the positive action and state pseudo labels (A) implicitly serve as negatives for all other categories via the softmax output function. Therefore, the model utilizes the power of these cross-task negatives implicitly. For the multi-classifier architecture (option~{II.}), we explicitly use random frames from videos of different categories in a batch as action and state negatives.

\paragraphcustom{End-to-end finetuning.}
Prior works \cite{Alayrac16ObjectStates,soucek2022lookforthechange} use frozen pre-extracted features as input for their classifiers. While training the whole model could yield substantial performance gains, doing so requires enormous computational and memory resources due to the very long video inputs, as observed by Wu \etal{} \cite{wu2019long}. 
However, our algorithm is sparse and returns non-zero gradients only for a fixed number of frames per video. This fact caused by the nature of our constraints leads us to the following two-stage algorithm that enables end-to-end training. We first run inference on all videos in a batch to identify a fixed number of positive and negative frames given by the constraints (this is akin to mining positives and hard-negative mining). Then we perform a forward and backward pass only on these identified video frames. On the ChangeIt dataset, our method back-propagates only through less than ten percent of the video frames (\ie{} 25 frames per video at the resolution of 1 fps), effectively requiring only a tenth of the memory to run compared to the naive approach.

\section{Experiments}
\label{sec:experiments}

In this section, we first validate our contributions using a set of carefully chosen ablations. 
Second, we quantitatively compare our model with the current state-of-the-art on the large-scale ChangeIt~\cite{soucek2022lookforthechange} and COIN datasets~\cite{tang2019coin}. Third, we evaluate the zero-shot capabilities of our model on egocentric EPIC-KITCHENS~\cite{damen2018scaling} and Ego4D~\cite{grauman2021ego4d} datasets. Finally, we show qualitative results and discuss the main limitation of our method. 
We refer the reader to \ARXIVversion{Appendix~\ref{supmat:implementation} and \ref{supmat:eval}}{the supplementary material} for implementation and evaluation details as well as additional results and their discussion.

\begin{table}
  \centering
  {\footnotesize
  \begin{tabular}{c@{~~~}c@{~~~}c@{~~~}r@{~}c@{~}l@{~}c@{~}c@{~}c@{~}c@{~~~}c@{~~~}c}
    \toprule
    & \rotatebox{90}{\footnotesize architecture (fig. \ref{fig:architectures})} & \rotatebox{90}{\footnotesize\makecell[l]{\# of state +  \vspace{-1.5pt}\\ action heads}} & \multicolumn{3}{c}{\rotatebox{90}{\footnotesize\makecell[l]{\vspace{-5pt}\\ \# of outputs per \vspace{-1.5pt}\\ state / action head}}} & \rotatebox{90}{\footnotesize\makecell[l]{action background \vspace{-1.5pt}\\ negatives (fig. \ref{fig:targets}B)}} & \rotatebox{90}{\footnotesize\makecell[l]{cross-task \vspace{-1.5pt}\\ negatives (fig. \ref{fig:targets}E)}} & \rotatebox{90}{\footnotesize\makecell[l]{state as action \vspace{-1.5pt}\\ negatives (fig. \ref{fig:targets}C)}} & \rotatebox{90}{\footnotesize\makecell[l]{action as state \vspace{-1.5pt}\\ negatives (fig. \ref{fig:targets}D)}} & \rotatebox{90}{\footnotesize state prec.}& \rotatebox{90}{\footnotesize action prec.} \\
    \midrule
    \cite{soucek2022lookforthechange} & I & 1+1\textsuperscript{$\dagger$} & 2&/&1 & \cmark & \xmark & \xmark & \xmark & 0.35 & 0.68 \\
    \midrule
    \textbf{(a)} & II & N+N & 2&/&1 & \cmark & \xmark & \xmark & \xmark & 0.36 & 0.68 \\
    \textbf{(b)} & II & N+N & 3&/&1 & \cmark & \cmark & \xmark & \xmark & 0.42 & 0.73 \\
    \textbf{(c)} & II & N+N & 3&/&1 & \cmark & \cmark & \cmark & \xmark & 0.41 & 0.73 \\
    \textbf{(d)} & II & N+N & 3&/&1 & \cmark & \cmark & \cmark & \cmark & 0.41 & 0.73 \\
    \midrule
    \textbf{(e)} & III& 1   & \multicolumn{3}{c}{3N} & \xmark & \cmark & \xmark & \xmark & 0.37 & 0.67\\
    \textbf{(f)} & IV & 1+1 & 2N&/&N+1   & \cmark & \cmark & \xmark & \xmark & 0.43 & 0.72\\
    \textbf{(g)} & IV & 1+1 & 2N&/&N+1   & \cmark & \cmark & \cmark & \xmark & 0.42 & 0.75\\
    \textbf{(h)} & IV & 1+1 & 2N+1&/&N+1 & \cmark & \cmark & \cmark & \cmark & \textbf{0.47} & \textbf{0.77}\\
    \bottomrule
    \multicolumn{12}{c}{\textsuperscript{$\dagger$} One model per category.}\\
  \end{tabular}
  }
  \vspace{-0.5em}
  \caption{
  \textbf{Ablating different multi-task model architectures and strategies for constructing self-supervised pseudo labels.}
  For comparison, we also add the single-task setup of \cite{soucek2022lookforthechange}.
  }
  \label{tab:changeitAblation}
  \vspace{-0.5em}
  \vspace{-0.5em}
\end{table}

\paragraphcustom{Ablations.}
In this section, we investigate the proposed solutions for our task of automatically discovering state-modifying actions and corresponding object states jointly for all interaction categories. We train all architecture variants described in Section~\ref{sec:model_architecture} with frozen backbone (for simplicity) and employ the sampling strategies from Section~\ref{ref:model_constraints}. We run all experiments on the whole ChangeIt dataset while using the standard evaluation procedure used in prior work~\cite{soucek2022lookforthechange} and described in \ARXIVversion{Appendix~\ref{supmat:eval}}{the supplementary material}. Our findings are summarized in Table~\ref{tab:changeitAblation} and detailed below.

A simple extension of the prior work -- a model with $N$ state and $N$ action classifier heads, labeled~\textbf{(a)} in Table \ref{tab:changeitAblation} -- yields comparable results to the performance of the baseline approach of~\cite{soucek2022lookforthechange}. We observe a significant improvement when the cross-task negatives for both states and actions are added into the set of pseudo labels \textbf{(b)}.
Inspired by these results, we test a single-head joint-classifier model (Fig. \ref{fig:architectures} (III)), in Table \ref{tab:changeitAblation} denoted as \textbf{(e)}, which utilizes the cross-task negatives implicitly.
However, the performance is similar to the baseline classifier \textbf{(a)} due to the lack of explicit action background negatives.
Therefore, we train the two-head joint-classifier model (Fig. \ref{fig:architectures} (IV)), where a single background action class is added, with the action background negatives \textbf{(f)}. 
This set-up achieves similar performance as the multi-classifier model of \textbf{(b)}. 

We investigate adding other forms of negative pseudo labels as described in Section \ref{ref:model_constraints}. Specifically, we use the states as action negatives (Fig. \ref{fig:targets} (C)), in Table \ref{tab:changeitAblation} variants \textbf{(c)} and \textbf{(g)}, and the actions as state negatives (Fig. \ref{fig:targets} (D)), in Table \ref{tab:changeitAblation} variants \textbf{(d)} and \textbf{(h)}.
We observe significant performance improvements when adding both sets of pseudo-labels to the two-head joint-classifier model, variants \textbf{(g)} and \textbf{(h)}. However, the models \textbf{(a)} to \textbf{(d)} with $N$ separate classifiers do not improve.
We hypothesize these models with $N$ background classes have too much flexibility and overfit to the presented pseudo labels. Therefore, our final model is the variant \textbf{(h)}, the two-head joint-classifier model, consisting of two heads with $2N+1$ and $N+1$ outputs, \ie{} the architecture IV from Figure~\ref{fig:architectures}. The first head predicts the likelihood of $N$ initial states, $N$ end states, and the background class; the second head predicts the likelihood of $N$ actions and the background class. During training, this model utilizes all presented pseudo label types (Fig. \ref{fig:targets} (A)-(E)) making the best use of cross-task parameter sharing and the self-supervised in-video constraints.

\begin{table}
  \centering
  {\small
  \begin{tabular}{l@{~~~}c@{~~}c|@{~~}c@{~~}}
    \toprule
    \multirow{2}{*}{Method} & \multicolumn{2}{c|@{~~}}{ChangeIt\cite{soucek2022lookforthechange}} & COIN \cite{tang2019coin} \\
     & {\footnotesize St prec.} & {\footnotesize Ac prec.} & {\footnotesize Ac prec.} \\
    \midrule
    Random                                               & 0.15 & 0.41 & 0.42 \\
    Merlot Reserve \cite{zellers2022merlot}              & 0.27 & 0.57 & 0.69 \\
    CLIP ViT-L/14 \cite{clip}                            & 0.30 & \underline{0.63} & 0.65 \\
    VideoCLIP \cite{xu2021videoclip}                     & \underline{0.33} & 0.59 & \underline{0.72} \\
    Alayrac \etal{} \cite{Alayrac16ObjectStates}         & 0.30 & 0.59 & 0.57 \\
    Look for the Change \cite{soucek2022lookforthechange}& \underline{0.35} & \underline{0.68} & \underline{0.73} \\
    \midrule
    \textbf{Ours} (backbone from\cite{soucek2022lookforthechange}) & 0.47 & 0.77 & 0.79 \\
    \textbf{Ours} (ViT-L/14 frozen)                      & 0.47 & 0.75 & 0.77 \\
    \textbf{Ours} (ViT-L/14 finetuned)                   & \textbf{0.49} & \textbf{0.80} & \textbf{0.83} \\
    \bottomrule
  \end{tabular}
  }
  \vspace{-0.5em}
  \caption{Comparison with the state-of-the-art.}
  \label{tab:changeitSOTA}
  \vspace{-0.5em}
  \vspace{-0.5em}
\end{table}

\begin{table}
  \centering
  {\small
  \begin{tabular}{l@{~~~}c@{~~}c|c@{~~}c}
    \toprule
    \multirow{2}{*}{Method} & \multicolumn{2}{c|}{EPIC-K.\cite{damen2018scaling}} & \multicolumn{2}{c}{Ego4D \cite{grauman2021ego4d}} \\
     & {\footnotesize St mAP} & {\footnotesize Ac mAP} & {\footnotesize St mAP} & {\footnotesize Ac mAP}  \\
    \midrule
    Random                                               & 0.09 & 0.07 & 0.13 & 0.12 \\
    Merlot Reserve \cite{zellers2022merlot}              & \underline{0.31} & 0.36 & \underline{0.25} & 0.45 \\
    CLIP ViT-L/14 \cite{clip}                            & 0.23 & 0.35 & 0.23 & 0.42 \\
    VideoCLIP \cite{xu2021videoclip}                     & 0.25 & \underline{0.44} & 0.23 & \textbf{0.49} \\
    Look for the Ch. \cite{soucek2022lookforthechange}\textsuperscript{$\dagger$}  & 0.12 & 0.15 & 0.20 & 0.17 \\
    \textbf{Ours} (ViT-L/14)\textsuperscript{$\dagger$}  & \textbf{0.38} & \textbf{0.51} & \textbf{0.37} & \underline{0.48} \\
    \bottomrule
    \multicolumn{5}{c}{\textsuperscript{$\dagger$} Trained on the ChangeIt dataset, zero-shot evaluation.}\\
  \end{tabular}
  }
  \vspace{-0.5em}
  \caption{Zero-shot retrieval of object states and state-modifying actions on subsets of the EPIC-KITCHENS \cite{damen2018scaling} and the Ego4D~\cite{grauman2021ego4d} datasets.}
  \label{tab:egocentric}
  \vspace{-0.5em}
  \vspace{-0.5em}
\end{table}

\begin{figure*}
  \centering
  \includegraphics[width=1.0\linewidth]{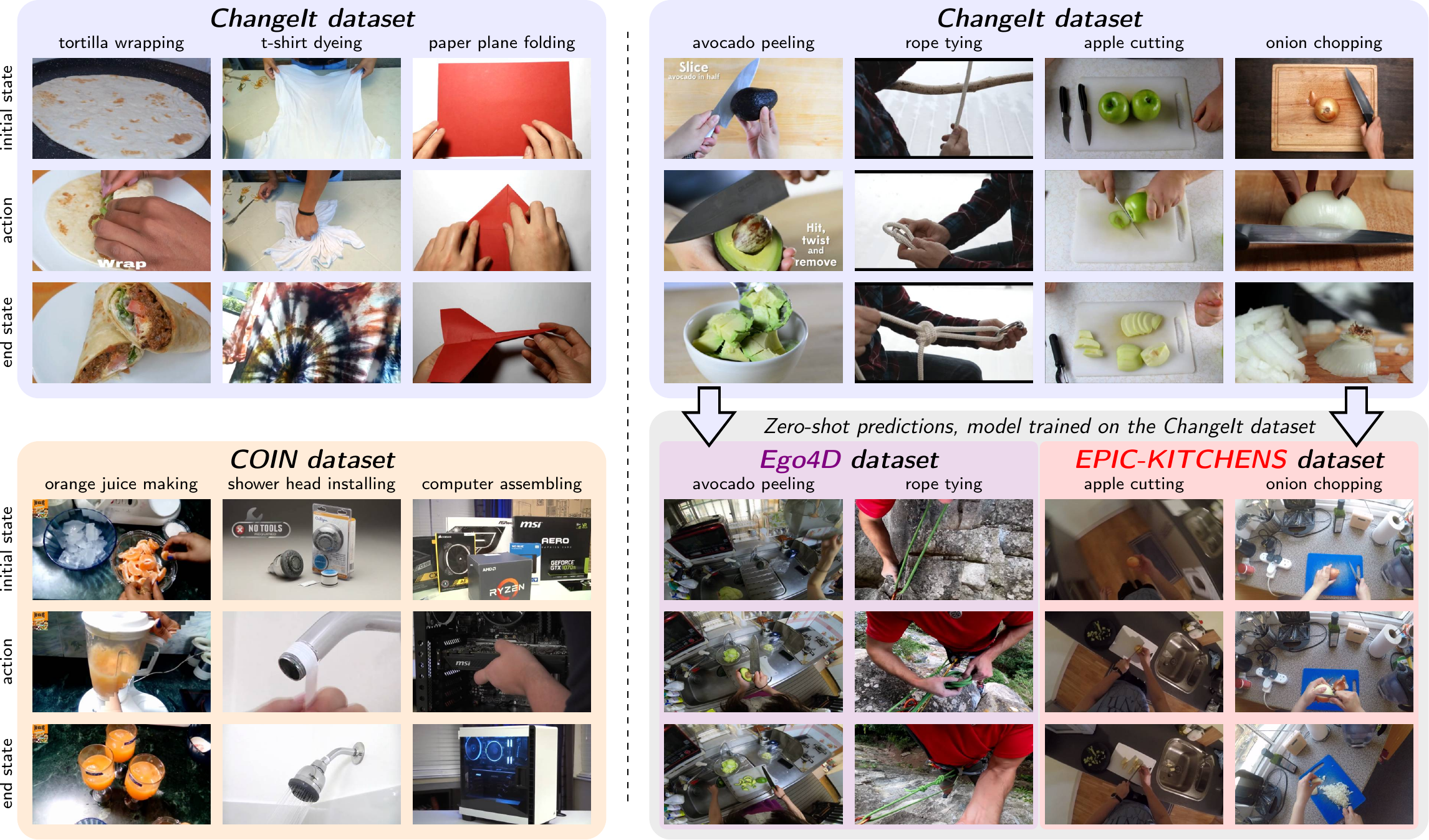}
  \vspace{-0.5em}
  \vspace{-0.5em}
  \caption{\textbf{Discovered object states and actions.} Each column shows three frames corresponding to (from the top) the initial state $\to$ state-modifying action $\to$ the end state. {\bf Left:} Different interaction categories from the ChangeIt and the COIN datasets showing the~diversity of discovered states and actions. {\bf Right:} The model learned on the ChangeIt dataset can be used to detect the learned interactions in egocentric videos of the Ego4D and the EPIC-KITCHENS datasets in a zero-shot setup. See \ARXIVversion{Appendix~\ref{supmat:qualitativeResults}}{the supplementary material} for additional examples.}
  \label{fig:qualitativeResults}
  \vspace{-0.5em}
  \vspace{-0.5em}
\end{figure*}

\paragraphcustom{Comparison with the state-of-the-art.}
We compare our method to the current state-of-the-art for state and action discovery -- Look for the Change \cite{soucek2022lookforthechange}. It introduced the causal ordering constraint as a method to bootstrap pseudo labels for a neural network classifier. However, this method trains a separate classifier per interaction category and cannot distinguish unrelated videos without the object of interest because the model lacks a \textit{background} class. Further, we compare our method to the work of Alayrac \etal{} \cite{Alayrac16ObjectStates} as well as various image-based \cite{clip} and video-based \cite{xu2021videoclip,zellers2022merlot} zero-shot methods. We provide evaluation details for all the methods in \ARXIVversion{Appendix~\ref{supmat:eval}}{the supplementary material}. 

We train our model with ViT-L/14 CLIP \cite{clip} backbone in an end-to-end manner as well as with the frozen backbones used by \cite{soucek2022lookforthechange} on the large-scale uncurated ChangeIt dataset and the curated COIN dataset. Results are shown in Table \ref{tab:changeitSOTA}.
Our model trained end-to-end outperforms the current state-of-the-art on the ChangeIt dataset by 14 percentage points in state precision and 12 percentage points in action precision, which corresponds to 40\% and 18\% relative improvements, respectively, demonstrating the benefits of our joint multi-task self-supervised learning setup. 
Significant improvements are also obtained on the COIN dataset, even though the dataset has fewer videos per category compared to the ChangeIt dataset making it challenging to discover the correct actions.
Further, we show that end-to-end finetuning brings a substantial performance boost compared to the variant without finetuning and the frozen backbones used by \cite{soucek2022lookforthechange} while using only a single image backbone.

\paragraphcustom{Zero-shot generalization to egocentric videos.}
Even though instructional videos on the web are rarely in egocentric view, having a strong model for egocentric videos is relevant for many applications \cite{grauman2021ego4d,damen2018scaling}. Therefore, we also compare our method on a set of videos from two egocentric datasets -- EPIC-KITCHENS \cite{damen2018scaling} and Ego4D \cite{grauman2021ego4d}. The datasets do not contain temporal state and action annotation; therefore, we manually annotate 23 hours of video depicting the object interactions present in the ChangeIt dataset. We test our model trained on the noisy uncurated web videos of the ChangeIt dataset on these egocentric videos. In Table \ref{tab:egocentric}, we show that our model outperforms previous works as well as image-based and most video-based zero-shot methods by a significant margin, demonstrating both strong performance and robustness of our model on out-of-distribution (here egocentric) data.

\paragraphcustom{Qualitative results.}
We retrieve the videos with the highest classification score $p(c|v)$ for a given interaction category~$c$ and show them in Figure~\ref{fig:qualitativeResults}. 
These results show that our model can: \textbf{(i)} retrieve the correct videos, ignoring videos without the object of interest; \textbf{(ii)} correctly temporally localize the initial state, the state-changing action, and the end state; and \textbf{(iii)} be successfully applied in zero-shot setup without any training to egocentric videos. In \ARXIVversion{Appendix~\ref{supmat:qualitativeResults}}{the supplementary material}, we also show the model can: \textbf{(iv)} filter out unrelated videos that pollute the search results during the dataset acquisition (videos with the classification score close to zero $p(c|v)\approx 0$); and \textbf{(v)} be consistent across videos, \ie{} the predicted state or action frame displays the object in the same visual state for every retrieved video. We show more qualitative results including failure cases in \ARXIVversion{Appendix~\ref{supmat:qualitativeResults}}{the supplementary material}.

\paragraphcustom{Limitations and societal impact.}
While we believe our model can bring new insights into unlabeled or weakly-labeled data, it is still prone to biases stemming from both the availability and the quality of the data. As expected, we observe that our model struggles with interaction categories having fewer and lower-quality videos. However, not all interactions are represented equally, especially if the datasets are gathered (semi)automatically. In such cases, our model can perform poorly on underrepresented interactions or outright ignore some manifestations of a given interaction if there are other more prevalent manifestations -- possibly ignoring interactions performed by minorities.
On the positive side, our work makes a step towards learning causal interactions from videos -- one of the key open problems in embodied video understanding with potential applications in large-scale learning of reward functions for visually guided robotics. For other limitations, please see \ARXIVversion{Appendix~\ref{supmat:impact}}{the supplementary material}.

\section{Conclusion}
We have developed a new multi-task method for jointly learning object states and state-modifying actions with minimal supervision from long, uncurated web videos. Our approach introduces a multi-task architecture and a pseudo-label sampling strategy, which are both key to achieving state-of-the-art results on this problem. 
We have shown our multi-task method significantly outperforms single-task approaches from previous work on the large-scale ChangeIt dataset. We have also demonstrated our learned models can be used in a zero-shot manner on egocentric datasets such as Ego4D despite training mostly on third-person views. This opens up the possibility of truly Internet-scale learning about changes of object states caused by human actions.

\paragraphcustom{Acknowledgements.}
This work was partly supported by the European Regional Development Fund under the project IMPACT (reg. no. CZ.02.1.01/0.0/0.0/15\_003/0000468), the Ministry of Education, Youth and Sports of the Czech Republic through the e-INFRA CZ (ID:90140), the French government under management of Agence Nationale de la Recherche as part of the ``Investissements d'avenir'' program, reference ANR19-P3IA-0001 (PRAIRIE 3IA Institute), and Louis Vuitton ENS Chair on Artificial Intelligence.

{\small
\bibliographystyle{ieee_fullname}
\bibliography{egbib}
\flushcolsend
}

\ARXIVversion{\newpage\appendix\ARXIVversion{\section*{Appendix}}{\section*{Overview}}
\ARXIVversion{We}{In this supplementary material, we} provide the implementation details of our model in Section \ref{supmat:implementation}. 
Additional details of the evaluation setup are given in Section \ref{supmat:eval}. In Section \ref{supmat:classification} we test our model's classification performance on the ChangeIt dataset. In Section \ref{supmat:qualitativeResults} we show additional qualitative results and describe the limitations of our method. Then we introduce our other supplementary content in Section \ref{supmat:video}. Finally, in Section \ref{supmat:impact} we discuss the limitations of our work.

\section{Implementation details}\label{supmat:implementation}
\paragraphcustomWOvspace{Model architecture.} Our final model consists of a ViT-L/14 image backbone initialized with CLIP pre-trained weights~\cite{clip}. We feed the 1024-dimensional feature representation before the joint-space projection into two randomly initialized MLP classifiers for states and actions. Both MLPs consist of one hidden layer of dimension 4096 with the ReLU activation function. The input to the model is the video frames at a temporal resolution of one frame per second. Each frame is processed independently of the adjacent frames. While we acknowledge the independent processing of each frame is suboptimal, it is used due to the favorable computational resource requirements. We use standard image pre-processing and augmentation techniques, \ie, we apply random flipping, rotation, clipping, blurring, and color changes, such as brightness, saturation, and hue. We apply the exact same augmentation to all frames of a given video.

For the ablation experiments, we used the same setup as in~\cite{soucek2022lookforthechange}, \ie, we used pre-extracted feature vectors instead of extracting the feature vectors during training and back-propagating into the backbones to save time and computational resources.

\paragraphcustom{Hyperparameters.}
We optimize the MLP layers using stochastic gradient descent with a momentum of $0.9$ and $L_2$ penalty of $0.001$. The backbone is optimized using AdamW optimizer without penalty~\cite{loshchilov2018decoupled}. For the learning rate, we use a cosine schedule with a linear warmup of five epochs; the base learning rate for the MLPs is $10^{-4}$ and $10^{-5}$ for the backbone. We train our model with a batch size of 64 videos randomly sampled across all dataset categories and distributed over 32 A100 GPUs. The training for 50 epochs takes two days.

For sampling pseudo labels, we use a neighborhood~$\delta$ of two seconds (\ie, two frames at our temporal resolution of 1~fps) and the distance for the action background negatives~$\delta'$ is 60 seconds. The $\omega(v,q)$ weight in Equation~\ARXIVversion{\eqref{eq:loss}}{(3) in the main paper} balances the state and action classifier losses by multiplying terms for the action classifier (\ie, if $q=a$ or $q=\emptyset_a$) by $0.2$. For the ChangeIt dataset, we also weighed each loss term by the noise adaptive weight from \cite{soucek2022lookforthechange}. However, we observed only a small difference in the performance.

\section{Evaluation details}\label{supmat:eval}
\paragraphcustomWOvspace{ChangeIt dataset.}
The ChangeIt dataset \cite{soucek2022lookforthechange} contains more than 34,000 in-the-wild YouTube videos with an average length of 4.6 minutes in 44 various interaction categories, such as \textit{apple cutting} and \textit{paper plane folding}. The videos are obtained automatically; therefore, a large portion of the videos is noisy or outright unrelated. We follow the evaluation protocol of the prior work~\cite{soucek2022lookforthechange} for computing state and action precision on the test set. We predict the temporal locations for each test video and its corresponding interaction category for the initial state, the action, and the end state and measure the precision for states and actions. We average the metrics over videos of each category and finally report the average over all categories. To be consistent with the prior work, we take model weights from the best-performing epoch and report their performance averaged over three runs, computed for each category independently.

\paragraphcustom{COIN dataset.}
The COIN dataset \cite{tang2019coin} contains a curated set of approximately 11 thousand Youtube videos, which is about one third of the size of the ChangeIt dataset, with an average video length of 2.4 minutes. The dataset contains 180 object interaction categories such as \textit{changing bike tires}. Various sub-actions such as \textit{remove the tire} or \textit{pump up the tire} are temporarily annotated in each video. As our task is to discover only a single temporal location, we merge all sub-actions into a single \textit{action} category. The action precision is then computed as in the case of the ChangeIt dataset. Because the COIN dataset contains only temporal action annotation, we only evaluate and report action precision. However, our method also estimates the temporal locations of object states in this data.

\paragraphcustom{Ego4D and EPIC-KITCHENS datasets.} The egocentric datasets Ego4D \cite{grauman2021ego4d} and EPIC-KITCHENS \cite{damen2018scaling} together contain close to four thousand hours of egocentric video. While both datasets contain object annotations, the datasets do not contain information about object states. Therefore, we gather from those dataset videos containing common objects appearing in the ChangeIt dataset, such as \textit{an onion} and \textit{a rope}. We annotate 60 videos totaling 23 hours with the corresponding action, initial-, and end-state temporal labels. As the egocentric videos are much longer with many more object and action occurrences, we predict probabilities for the action, the initial and the end state for each video frame and report the mean average precision (mAP). We compute average precision separately for the retrieval of the action, the initial and the end state in each video, and average it across the three classes and across all the videos. We do not train our model on the egocentric videos; we use a model trained on the ChangeIt dataset in a zero-shot setup to evaluate our generalization performance.

\paragraphcustom{Baseline details.}
The methods of Alayrac \etal{} \cite{Alayrac16ObjectStates} and Look for the Change~\cite{soucek2022lookforthechange} both directly output a single location of the action, the initial and the end state. The zero-shot methods, CLIP~\cite{clip}, Merlot Reserve~\cite{zellers2022merlot}, and VideoCLIP~\cite{xu2021videoclip}, return a similarity between a text description and an image or video segment. In our setup, we compute the cosine similarity between a state or action text description (the prompt) and every video frame (or video segment). For the ChangeIt and the COIN datasets, we use Equation~\ARXIVversion{\eqref{eq:argmax}}{(2) in the main paper} with these similarities to produce the final state and action locations. We use multiple text descriptions (prompts) both for the states and the actions and report the performance of the best-performing ones.

\section{Using our model for classification}\label{supmat:classification}
Our method has the ability to classify any video into the $N$ interaction categories or the background category if the video does not contain any of those $N$ categories. Therefore, we compare our method to several classification baselines on the ChangeIt dataset. We use the set of 667 videos with state and action per-frame annotations, where the video category label has been confirmed by a human annotator, as our test set. For the train set, we use the remainder of the videos with their automatically obtained noisy labels.
For our method, we compute the category for each test video as $\hat{c}=\argmax_c p(c|v)$. We compare our results with the zero-shot CLIP~\cite{clip}, VideoCLIP~\cite{xu2021videoclip}, and Merlot Reserve~\cite{zellers2022merlot} models, as well as a single layer model~\cite{scikit-learn} trained to predict the noisy video label using frame features extracted CLIP ViT-L/14~\cite{clip}, averaged over all video frames  (see below for details). We report the classification accuracy of our model and the baselines in Table~\ref{tab:changeitClassification}. Our method significantly outperforms the zero-shot models and beats the performance of the supervised single-layer model while not being directly trained to classify the videos.

Furthermore, we observe that, by detecting and localizing the action and the object states in a video, our method can filter out unrelated content of the video, such as people interactions, animations, intro screens, or end credits. Therefore, we also train a single layer model where each video $v$ with noisy interaction category label $l$ (used to query the search engine) is represented only by features of the frames corresponding to action and state locations $d_l^{a}(v)$, $d_l^{s_1}(v)$, $d_l^{s_2}(v)$ provided by our model and averaged together. On the test set videos, never before seen by the single layer classifier nor by our detection model, the classifier achieves almost perfect accuracy of 98\% (Table \ref{tab:changeitClassification}, last row). Even though such a model is not practical for classification as it requires the knowledge of the noisy interaction category label (used to query the search engine) for the test set videos, it shows the immense power of our state and action discovery method that can be utilized to filter results obtained from a search engine query in downstream applications such data cleaning or human-assisted data labeling.

\begin{table}
  \centering
  {\small
  \begin{tabular}{l@{~~~}c}
    \toprule
    Method & Acc. \\
    \midrule
    Random                                 & 0.02 \\
    VideoCLIP \cite{xu2021videoclip}       & 0.66 \\
    Merlot Reserve \cite{zellers2022merlot}& 0.68 \\
    CLIP ViT-L/14 \cite{clip}              & 0.82 \\
    Linear (averaged frame-level features) & 0.88 \\
    \textbf{Ours}                          & \textbf{0.90} \\
    \midrule
    Linear (model-selected frame-level feats.) & 0.98 \\
    \bottomrule
  \end{tabular}
  }
  \caption{Classification results on the ChangeIt dataset.}
  \label{tab:changeitClassification}
\end{table}

\begin{figure*}
  \centering
  \includegraphics[width=1.0\linewidth]{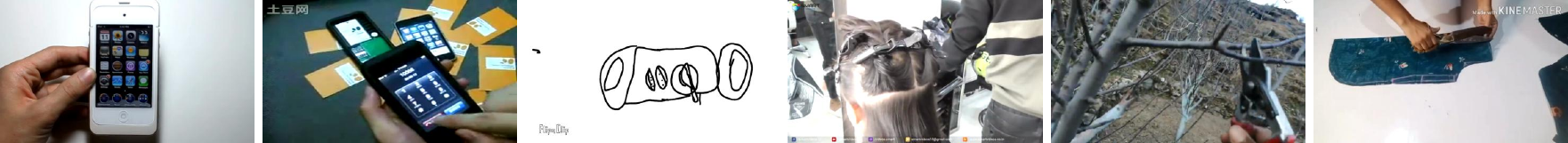}\\
  \vspace{0.15cm}
  \includegraphics[width=1.0\linewidth]{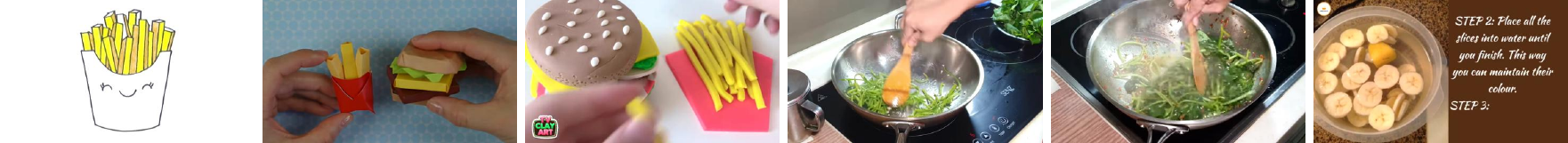}\\
  \caption{\textbf{Frames from videos of} \textit{apple cutting} (top) and \textit{potatoes frying} (bottom) \textbf{in the ChangeIt dataset classified as unrelated} by our model ($p(c|v)\approx 0$). Our model can detect noise in the dataset caused by language ambiguity. In the top row: phones manufactured by \textit{Apple} Inc., hair \textit{cutting}, \textit{apple} tree \textit{cutting}, \etc{} In the bottom row: sketch of \textit{fries}, \textit{fries} as plastic toys, beans or bananas \textit{frying}.}
  \label{fig:noisyVideos}
\end{figure*}

\paragraphcustom{Classification baselines.}
For the zero-shot models, we predict the category of a video as follows: First, we compute the cosine similarities of each video frame (or video segment) with text descriptions of the actions and states (the prompts). Then, we use a similar prediction scheme as for our method, \ie{} we use Equation~\ARXIVversion{\eqref{eq:max}}{(1) in the main paper} with the cosine similarity scores the category with the largest score is selected as the predicted one. For the single-layer model, we use CLIP ViT-L/14~\cite{clip} 1024-dimensional features before the joint-space projection layer extracted from video frames sampled at 1 fps and averaged together to produce a single 1024-dimensional vector. We implement the model using scikit-learn \cite{scikit-learn} and do a hyper-parameter sweep on the test set and report the best results. We observe fast overfitting to the noisy data; thus, the best hyper-parameters utilize strong regularization and early stopping.

\section{Additional qualitative results}\label{supmat:qualitativeResults}
Figures \ref{fig:qualitativeResultsSupmat}, \ref{fig:qualitativeResultsSupmat2}, and \ref{fig:qualitativeResultsSupmatCOIN} illustrate videos with the highest scores $p(c|v)$ for a given interaction category $c$ from the ChangeIt and COIN datasets. As mentioned in \ARXIVversion{Section~\ref{sec:experiments}}{the main paper}, our model can correctly retrieve relevant videos and temporally localize corresponding states and actions while being consistent across the videos of the interaction category. The model can also correctly localize the object states in videos of the COIN dataset, even though the dataset is action-centric and lacks object state annotations.
We observe that our model can correctly classify fine-grained interactions, such as \textit{tie tying} in Figure~\ref{fig:qualitativeResultsSupmat}.
Also, a clear example of the model's consistency is \textit{tortilla wrapping} interaction in Figure~\ref{fig:qualitativeResultsSupmat2} -- we can see the model always chooses a plain flat tortilla on a table for the initial state and a filled wrapped tortilla cut into halves as the end state. Additionally, in Figure~\ref{fig:qualitativeResultsSupmatEGO}, we show detected states and actions in egocentric videos of the Ego4D and EPIC-KITCHENS datasets, as done by our model trained only on the ChangeIt dataset and used in the zero-shot setup.

\paragraphcustom{Failure modes.}
Figure \ref{fig:qualitativeResultsErrors} shows failure modes of our method: \textbf{(i)} The method can select a different-than-expected consistent state or action, such as a frame of an already-flipped pancake instead of the flipping action (Figure \ref{fig:qualitativeResultsErrors}, \textit{pancake flipping}) or a frame of a cleaning brush instead of a dirty shoe as the initial state (Figure \ref{fig:qualitativeResultsErrors}, \textit{shoe cleaning}). 
\textbf{(ii)} Another failure mode corresponds to large variations of the visual appearance of an object, such as in the \textit{tree cutting} interaction (Figure \ref{fig:qualitativeResultsErrors}, \textit{tree cutting}) where only some branches, a standalone tree, or a forest can be cut. \textbf{(iii)} The method also struggles if there is only a minor visual difference between the initial and end states, such as a drilled hole in a piece of wood (Figure \ref{fig:qualitativeResultsErrors}, \textit{hole drilling}). However, the action, here \textit{drilling}, can be learned and localized correctly even if the states are not. 
This behaviour can be attributed to our use of \textit{`cross-modal'} negatives where the action classifier is forced to predict a background class at identified positions of initial and end states.

\paragraphcustom{Detecting unrelated videos.}
We examine videos with the classification score close to zero $p(c|v)\approx 0$, shown in Figure \ref{fig:noisyVideos}. We can see our model effectively filters out unrelated videos that pollute the search results during the dataset acquisition, such as videos of phones manufactured by \textit{Apple} Inc. in the interaction category \textit{apple cutting}.

\paragraphcustom{Visualizing related interactions in the shared representation.} 
We investigate the learned joint representation. In detail, we detect the temporal locations of the initial state, the end state, and the action for every video in selected categories and compute the Principal Component Analysis (PCA) of these detected frame representations. We use the output of the ViT backbone averaged over these three detected frames. We show both the original frozen CLIP feature representations as well as the finetuned feature representations produced by our model in Figure~\ref{fig:pca}. The figure shows our model can learn to separate small visual changes between related states and categories. The visualization also shows  that similar interactions, such as \textit{egg whisking} and \textit{cream whisking}, are close to each other in the feature space and dissimilar interactions are far apart demonstrating the learnt shared feature space.

\section{Additional limitations}\label{supmat:impact}
While our model significantly outperforms any prior work, it still inherits its limitations, mainly the fact that different interaction categories take a different number of steps to reach the best-performing parameters during learning.
After this point is reached, the model starts to overfit as it reinforces its predictions. 
Therefore, the best-performing model parameters may be different for different interaction categories. 
We acknowledge techniques such as multi-objective optimization \cite{lin2019pareto,Sener2018multiobjopt} could yield a possible solution, but we leave that for future work.

\section{\ARXIVversion{Code and video}{Supplementary video and code}}\label{supmat:video}
\ARXIVversion{We}{As part of the supplementary material, we} provide the code to train our model.\footnote{\label{fn:1}Available at \url{https://soczech.github.io/multi-task-object-states/}.} We also provide a video containing additional qualitative results on the ChangeIt, COIN, Ego4D, and EPIC-KITCHENS datasets.\footnoteref{fn:1} The video illustrates the difficulty of temporally localizing the object states and actions in long in-the-wild videos.

\begin{figure*}
  \centering

  \includegraphics[width=1.0\linewidth]{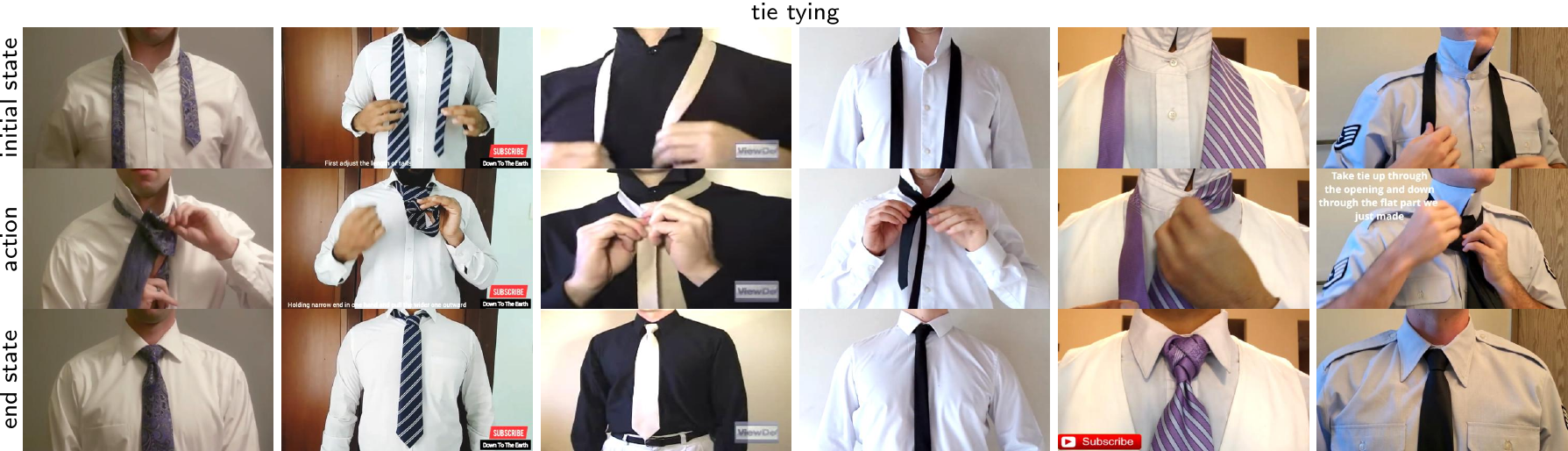}\\
  \vspace{0.15cm}
  \includegraphics[width=1.0\linewidth]{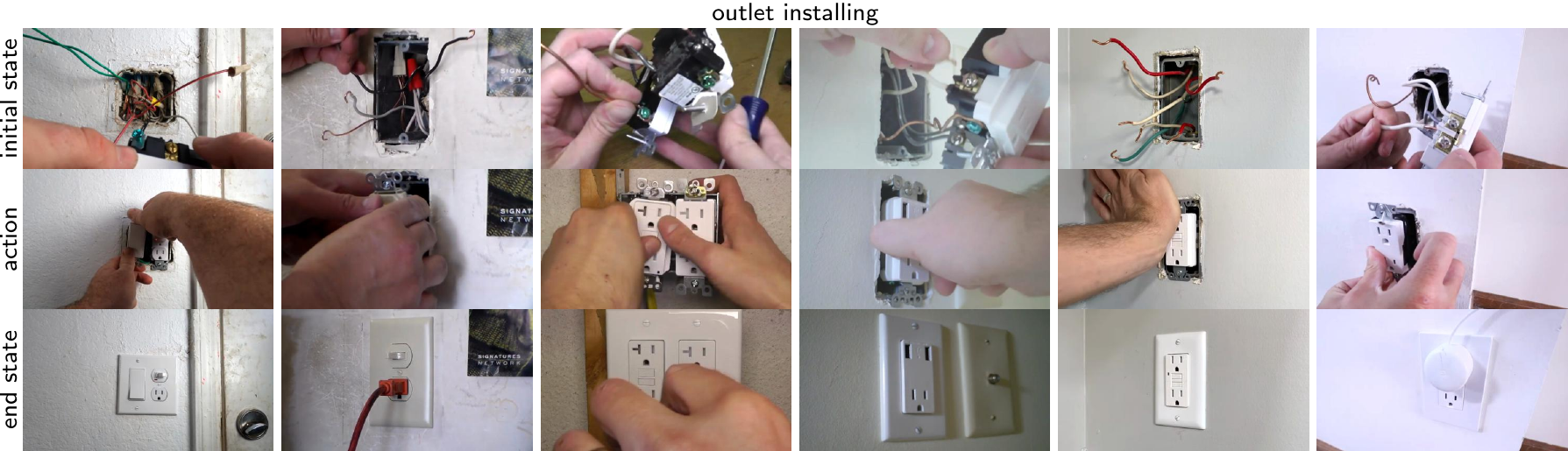}\\
  \vspace{0.15cm}
  \includegraphics[width=1.0\linewidth]{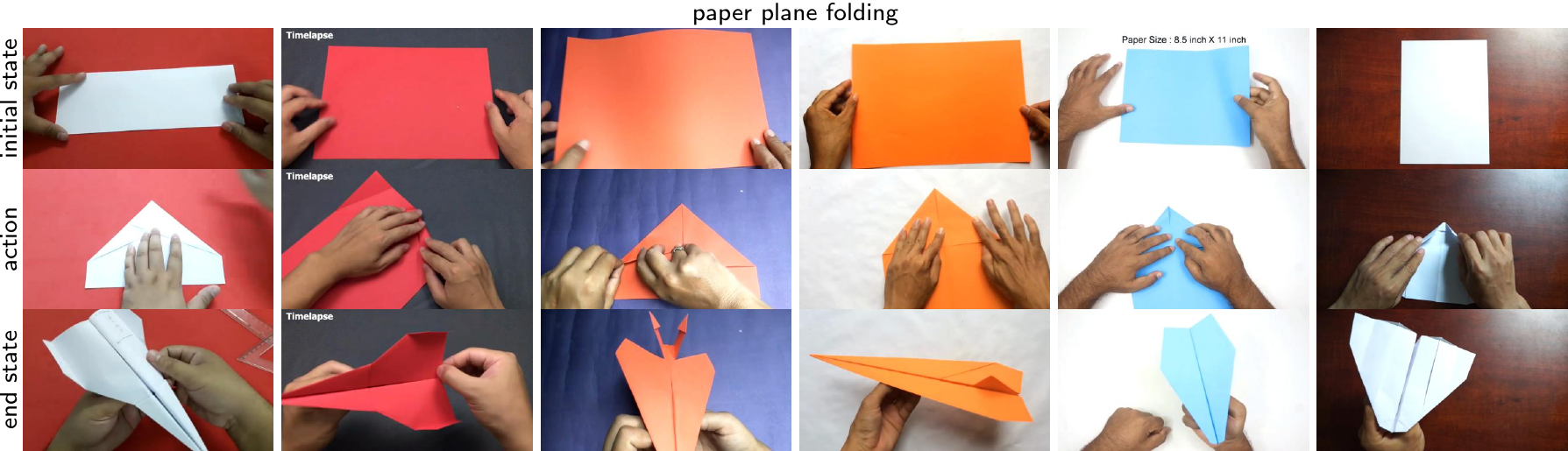}\\
  \vspace{0.15cm}
  \includegraphics[width=1.0\linewidth]{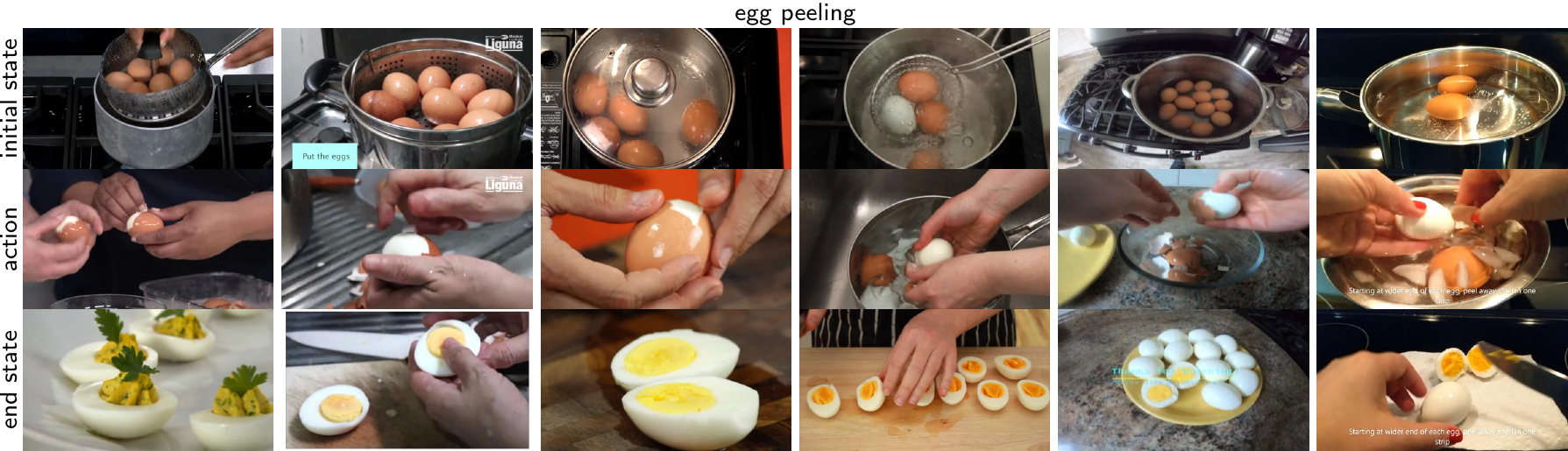}\\
  \vspace{-0.6em}
  \caption{\textbf{Discovered object states and actions in the \textit{ChangeIt} dataset.} Each plate shows examples of interactions demonstrating the visual consistency of the learned object states and actions. For each interaction category, each column shows three frames from one video, corresponding to the initial state  (top)  $\to$ state-modifying action (middle) $\to$ the end state (bottom).
  Note how our model can correctly detect fine-grained changes in appearance, \eg, in \textit{tie tying} interaction.
  }
  \label{fig:qualitativeResultsSupmat}
\end{figure*}

\begin{figure*}
  \centering
  \includegraphics[width=1.0\linewidth]{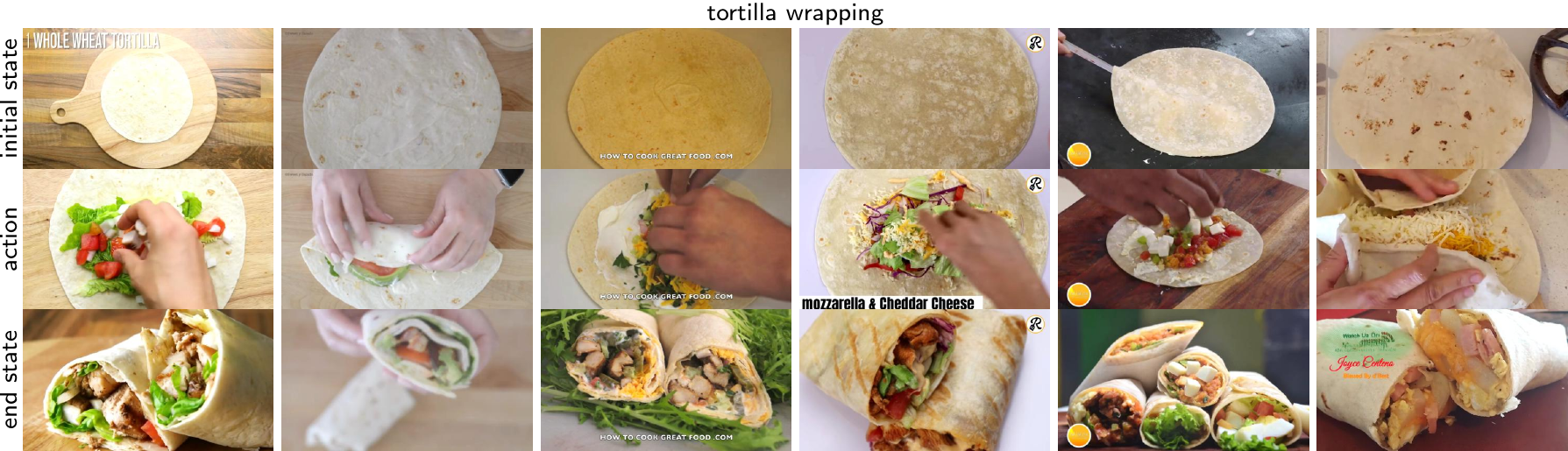}\\
  \vspace{0.1cm}
  \includegraphics[width=1.0\linewidth]{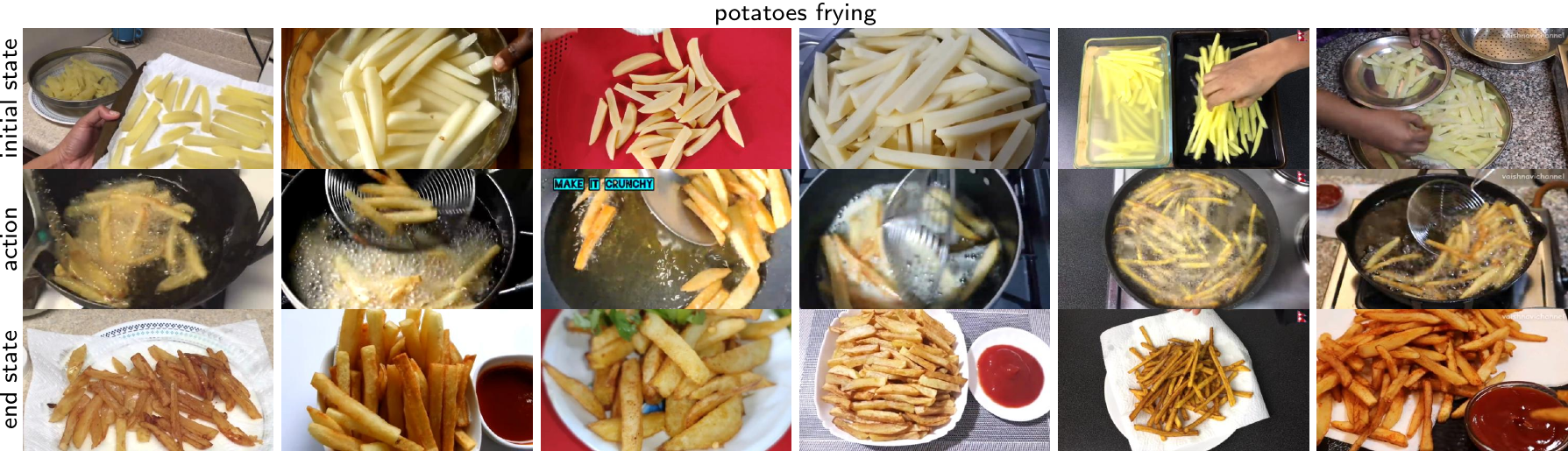}\\
  \vspace{-0.8em}
  \caption{\textbf{Additional examples of discovered object states and actions in the \textit{ChangeIt} dataset.} Each column shows three frames from one video, corresponding to the initial state  (top)  $\to$ state-modifying action (middle) $\to$ the end state (bottom). Note how the model consistently locates the same states and actions across all videos, \eg a plain flat tortilla on a table for the initial state of \textit{tortilla wrapping}. }
  \label{fig:qualitativeResultsSupmat2}
\end{figure*}

\begin{figure*}
  \vspace{-0.2cm}
  \centering
  \includegraphics[width=1.0\linewidth]{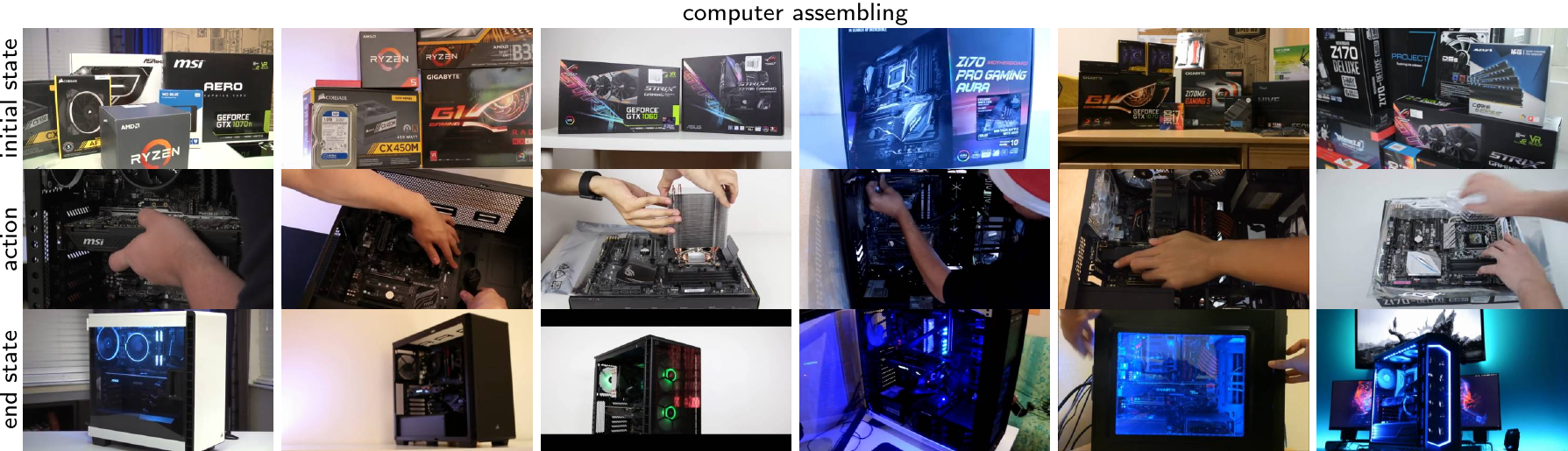}\\
  \vspace{0.1cm}
  \includegraphics[width=1.0\linewidth]{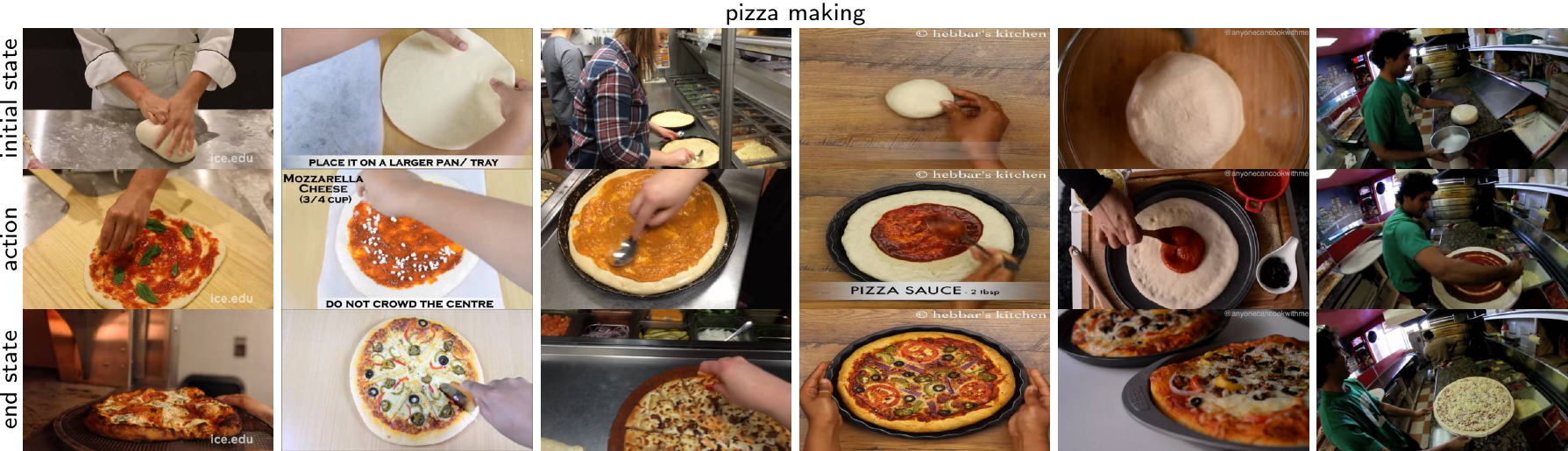}\\
  \vspace{-0.8em}
  \caption{\textbf{Examples of discovered object states and actions in the \textit{COIN} dataset.} Even though we do not evaluate state predictions on the COIN dataset due to the lack of object state annotations, our method correctly localizes the initial and end states in the COIN videos.}
  \label{fig:qualitativeResultsSupmatCOIN}
\end{figure*}

\begin{figure*}
  \centering
  \includegraphics[width=1.0\linewidth]{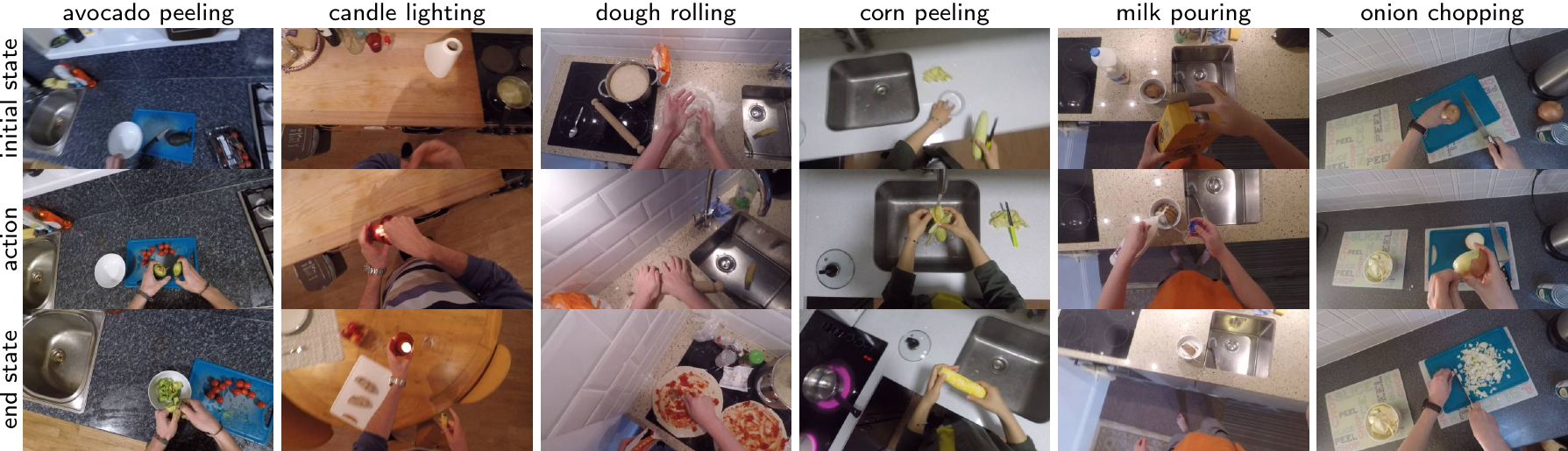}\\
  \includegraphics[width=1.0\linewidth]{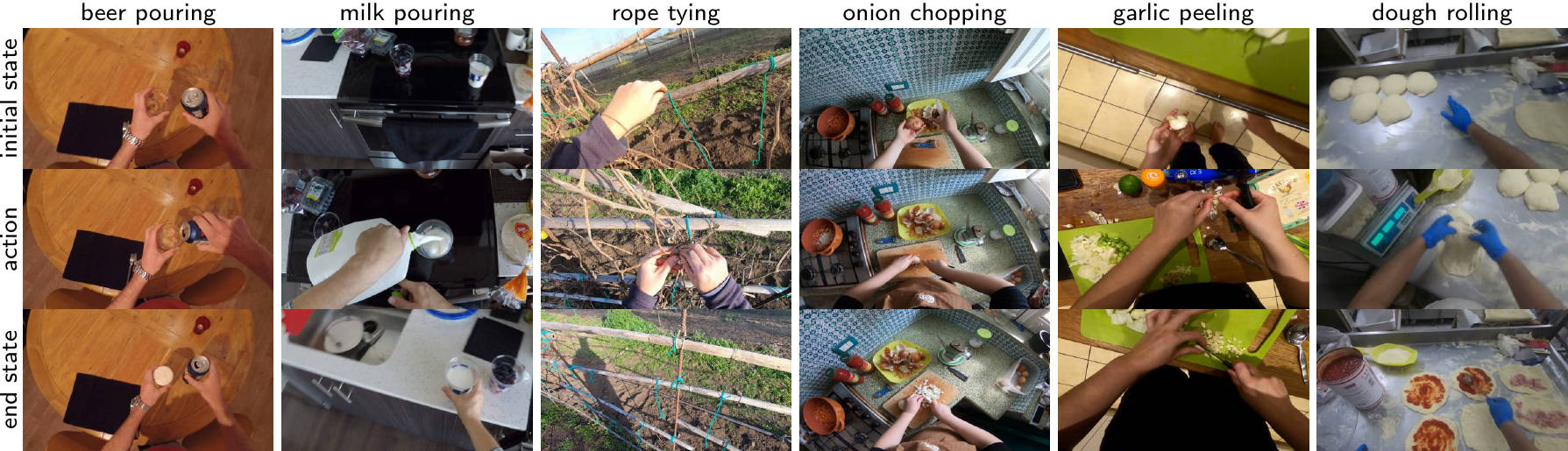}\\
  \vspace{-0.95em}
  \caption{\textbf{Examples of discovered object states and actions in egocentric videos of the \textit{Ego4D} and \textit{EPIC-KITCHENS} datasets.} Our model demonstrates its generalization abilities as it correctly temporally localizes the object states and the state-modifying actions in egocentric videos despite being trained only on noisy web videos of the ChangeIt dataset, captured mostly in third-person view.}
  \label{fig:qualitativeResultsSupmatEGO}
\end{figure*}

\begin{figure*}
  \vspace{-0.26cm}
  \centering
  \includegraphics[width=1.0\linewidth]{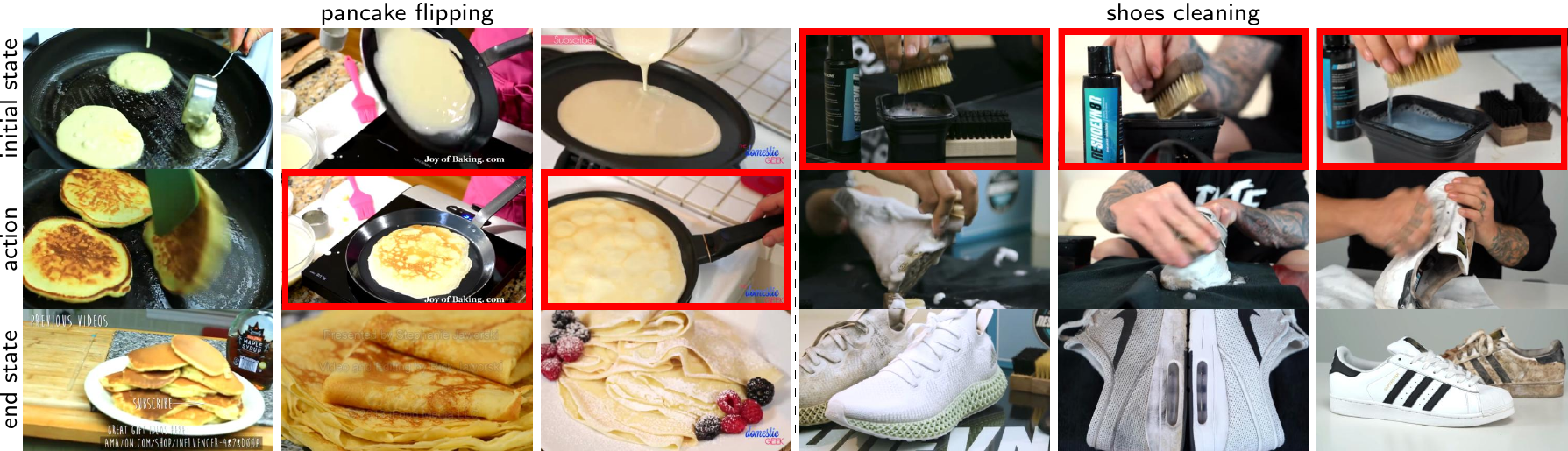}\\
  \includegraphics[width=1.0\linewidth]{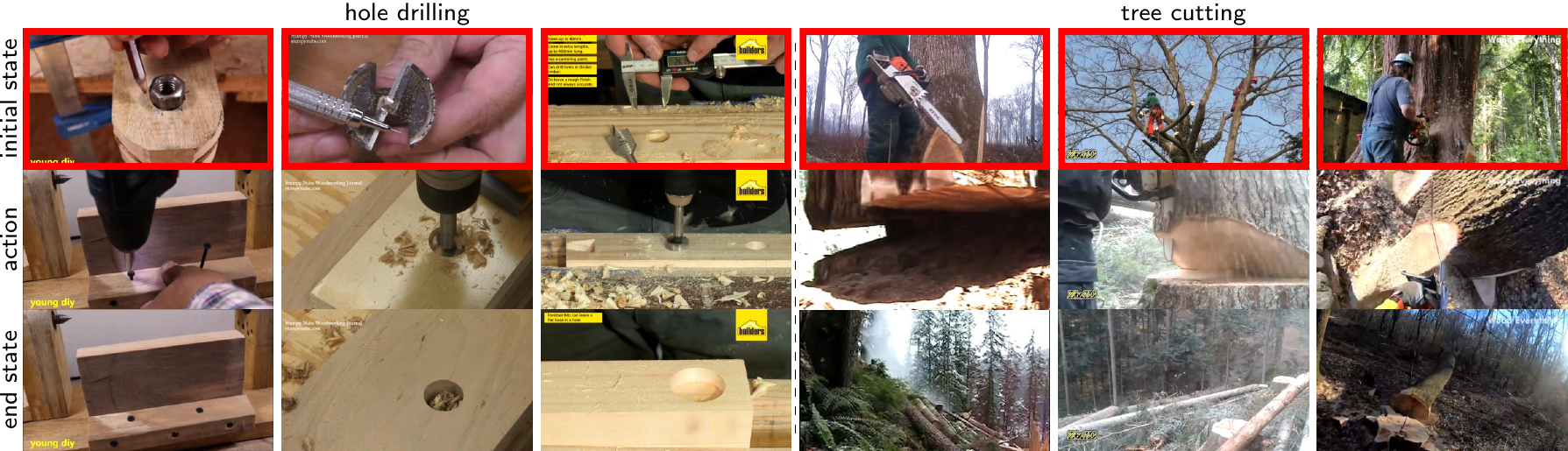}\\
  \vspace{-0.95em}
  \caption{\textbf{Examples of typical failure modes.} Selected failures are in red. The method can select a different-than-expected consistent state or action such as already flipped pancake for \textit{pancake flipping} or the brush for \textit{shoe cleaning}. It can also struggle if there is only a minor visual difference between the initial/end states as in \textit{hole drilling} or large variation in the appearance of the state as in \textit{tree cutting}.
  }
  
  \label{fig:qualitativeResultsErrors}
\end{figure*}

\begin{figure*}
  \centering
  \includegraphics[width=1.0\linewidth]{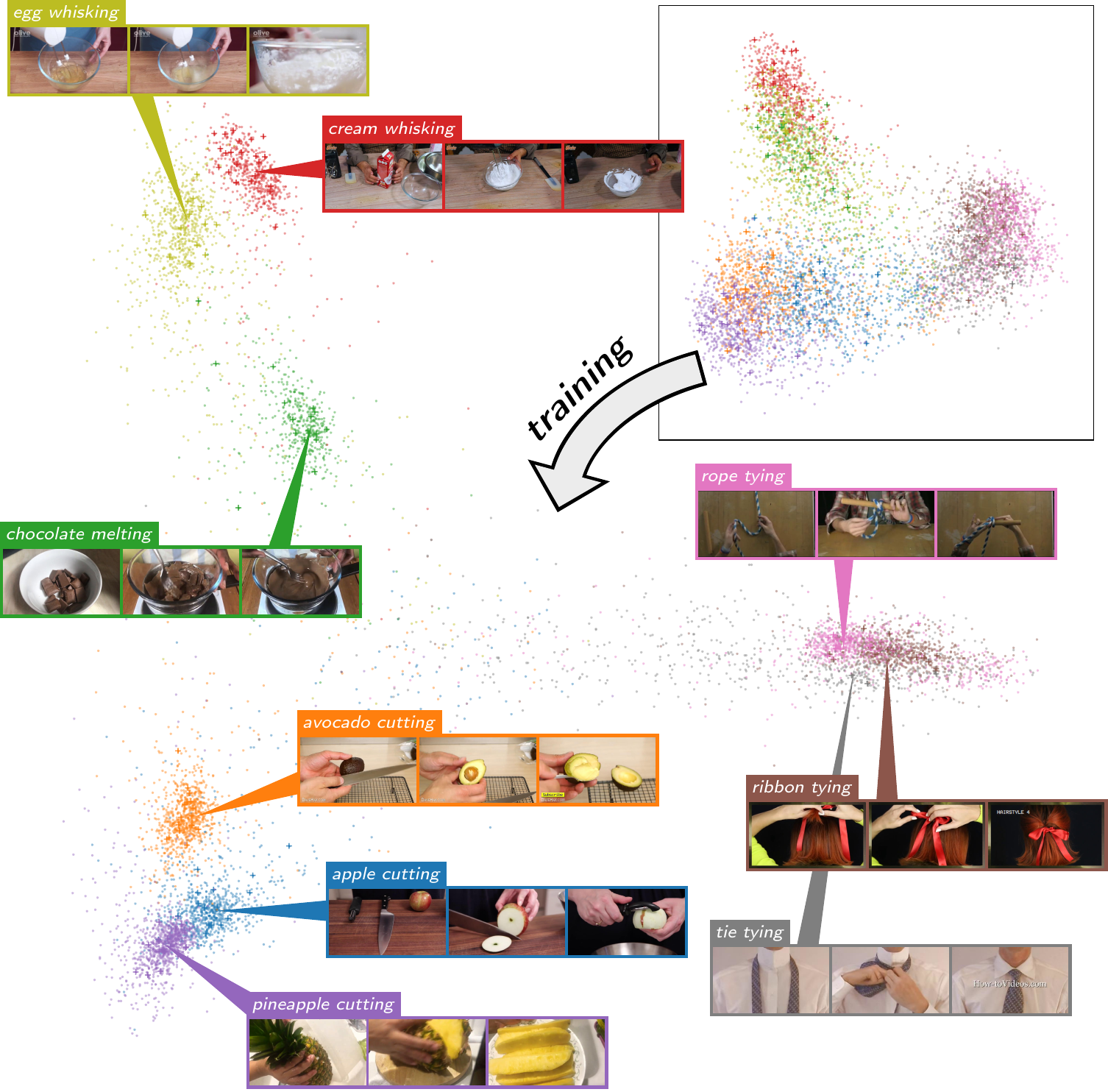}\\
  \caption{\textbf{Principal components of the learned feature space.} Each colored point represents a different video, where different colors denote different interaction categories such as \textit{rope tying} (pink) or \textit{ribbon tying} (brown), test videos are denoted by a plus sign. Each point representing a video is obtained in the following manner: We detect the temporal locations of the initial state, the end state, and the action for the video and compute the Principal Component Analysis (PCA) of the backbone feature representations of these detected frames. We show both our finetuned backbone representations (the main figure) and the original CLIP pre-trained representations (top right inset). Please note how related interactions, such as \textit{avocado cutting}, \textit{apple cutting}, and \textit{pineapple cutting}, are grouped in the feature space. In addition, similar interactions are also better separated from each other (\eg, \textit{egg whisking} and \textit{cream whisking}), compared to the frozen CLIP backbone features (top right inset). This is thanks to the multi-task constraints imposed by our approach.}
  \label{fig:pca}
\end{figure*}
}{}

\end{document}